\documentclass[journal]{IEEEtran}

\usepackage{booktabs}
\usepackage{graphicx}
\usepackage{changepage}

\newcommand{\PlotPathUnevenGround}{"figures"}

\title{Exoskeleton Knee Compliance Improves \\Gait Velocity and Stability \\in a Spinal Cord Injured User: A Case Report}

\author{Stefan O. Schrade, Giada Devittori, Christopher Awai Easthope, Camila Shirota, Olivier Lambercy and \\Roger Gassert% <-this % stops a space
	\thanks{S.O. Schrade, G. Devittori, C. Shirota, O. Lambercy, and R. Gassert are with the Rehabilitation Engineering Laboratory,
		ETH Zurich, 8008 Zurich, Switzerland   \hspace{3cm}
		{\tt\small \{relab.publications\}@hest.ethz.ch}}
	\thanks{C. Awai Easthop is with the Spinal Cord Injury Center, Balgrist University Hospital,
		University of Zurich, Forchstrasse 340, 8008, Zurich, Switzerland   \hspace{3cm}
		{\tt\small \{chris.schmidt\}@balgrist.ch}}
	}

\begin{document}

\maketitle

%TC:break Abstract
%the command above serves to have a word count for the abstract
\begin{abstract}
  Spinal cord injuries frequently impair the ability to walk. Powered lower limb exoskeletons offer a promising solution to restore walking ability. However, they are currently restricted to even ground. We hypothesized that compliant exoskeleton knees could decrease required effort to maneuver on uneven terrain, and increase gait velocity and stability. We describe a case study of a motor-complete spinal cord injury user (AIS~A, Th12) walking with a powered exoskeleton on even and uneven ground over multiple sessions after extensive training. Measurements with compliant or rigid exoskeleton knee joints were performed on three different days for each configuration. Body motion and crutch ground interaction forces were recorded to assess gait performance. We observed higher walking speeds with a compliant exoskeleton knee configuration (mean: 0.116~m/s on uneven and 0.145 m/s on even ground) compared to a rigid configuration (mean: 0.083~m/s and 0.100~m/s). Crutch force impulse was significantly reduced in the compliant configuration. Lastly, gait was more symmetric when the knee joints were compliant. In conclusion, compliant exoskeleton knee joints can help maneuver uneven ground faster and with less user effort than rigid joints. Based on our findings, exoskeleton designers should consider introducing compliance into their design to increase gait robustness and performance, and render exoskeletons more suitable for daily life use.
\end {abstract}
%TC:break main
%the command above serves to have a word count for the abstract

%\begin{keywords}
%gait analysis | walking speed | gait robustness | wearable robotics | force sensing crutches | lower limb
%\end{keywords}

%\begin{corrauthor}
%relab.publications\at hest.ethz.ch
%\end{corrauthor}

\section*{Introduction}

Powered lower limb exoskeletons are a promising solution to restore mobility after spinal cord injuries (SCI). These devices have recently become commercially available and are increasingly used by early adopters~\cite{Heinemann2018}. However, exoskeletons still cannot satisfyingly support walking on uneven ground despite their latent potential to do so. This restricts their use to mostly clinical environments and strongly limits home use. Leaving the structured clinical environment with even ground is one of the challenges to render exoskeletons true complements to wheelchairs for mobility, as the ability to walk over uneven ground is a feature that is highly prioritized by clinicians~\cite{Heinemann2018} and users alike. Together with limited walking speed~\cite{Louie2015} and significant costs~\cite{Jayaraman2016}, the current lack of versatility of existing exoskeletons may be the main reason why insurances are hesitant to reimburse exoskeletons for personal use.

Walking over uneven ground is a difficult task for exoskeleton users, as exoskeletons typically follow a pre-programmed trajectory with their legs~\cite{Tucker2015}. The highly geared,  rigid actuation systems do not allow deviations from this trajectory. Thus, uneven ground can lead to increased reliance on walking aids for balancing, reduced walking speed and decreased sense of safety.
Animals and humans, however, are easily able to cope with gait perturbations as has been demonstrated, e.g., in fowls running over a runway with a sudden drop in height~\cite{Daley2006}. Unimpaired humans adapt muscular control and modulate leg stiffness while running~\cite{Grimmer2008, Muller2010} and walking~\cite{Santuz2018} over uneven ground. Even throughout a single stride, the joint stiffness of the knee is known to be modulated~\cite{Pfeifer2014, Tucker2017}.

Research groups have taken inspiration from biological legs and suggested the use of compliant actuation designs, called variable stiffness actuation (VSA) systems, that allow varying joint stiffness. This family of actuators has been shown to allow efficient hopping~\cite{Vanderborght2009} and is similarly expected to bring benefits for walking. It is hypothesized that such mechanisms may also help to more robustly walk over uneven ground, rendering them a promising solution for use in powered leg exoskeletons~\cite{Junius2014, Bacek2017}.
However, benefits of such a system with impaired users have not been reported.
In general, walking over uneven ground with powered exoskeletons has not been thoroughly investigated, despite the large potential it offers to increase usability of this technology. One piece of work investigated walking of unimpaired users in an exoskeleton on a treadmill~\cite{Ugurlu2014}. During the experiment, obstacles were dropped on the treadmill and interaction forces at the handrail as well as torso pitch angular velocity were evaluated. The authors compared two different exoskeleton stiffness settings that they rendered via control, and found that angular velocity of the torso  and forces at the handrail were reduced if the exoskeleton was more compliant.

It remains unclear, however, how users with motor-complete spinal cord injury walking in an exoskeleton adapt to ground irregularities. Additionally, effects of joint stiffness on gait performance have not been investigated in this population. We employed the VariLeg exoskeleton~\cite{Schrade2018} to investigate the effect of exoskeleton knee joint compliance on walking over uneven ground in a case study. The VariLeg is equipped with variable stiffness actuators to modify the stiffness of the knee joints, and can also be used in conventional rigid actuation mode (no added compliance). The exoskeleton was used in two configurations: (i)~compliant knee joints and (ii)~rigid knee joints. With each exoskeleton configuration, the user walked over two ground types: (i)~even ground and (ii)~uneven ground. One user with motor-complete SCI was recruited and trained extensively to walk over the two ground types.  We tracked full body motion and measured the crutch ground interaction forces. We hypothesized that the compliant joints would facilitate maneuvering over uneven ground with the exoskeleton. We expected increased gait performance, reduced effort, and thus improved gait efficiency with a compliant knee joint configuration on uneven ground. We assumed this would manifest in reduced loading of the crutches and faster walking speed.

\section*{Materials and Methods}
\label{sec:Methods}

\subsection*{The VariLeg Exoskeleton}
\begin{figure*}
   % \centering
    \includegraphics[width=0.965 \textwidth]{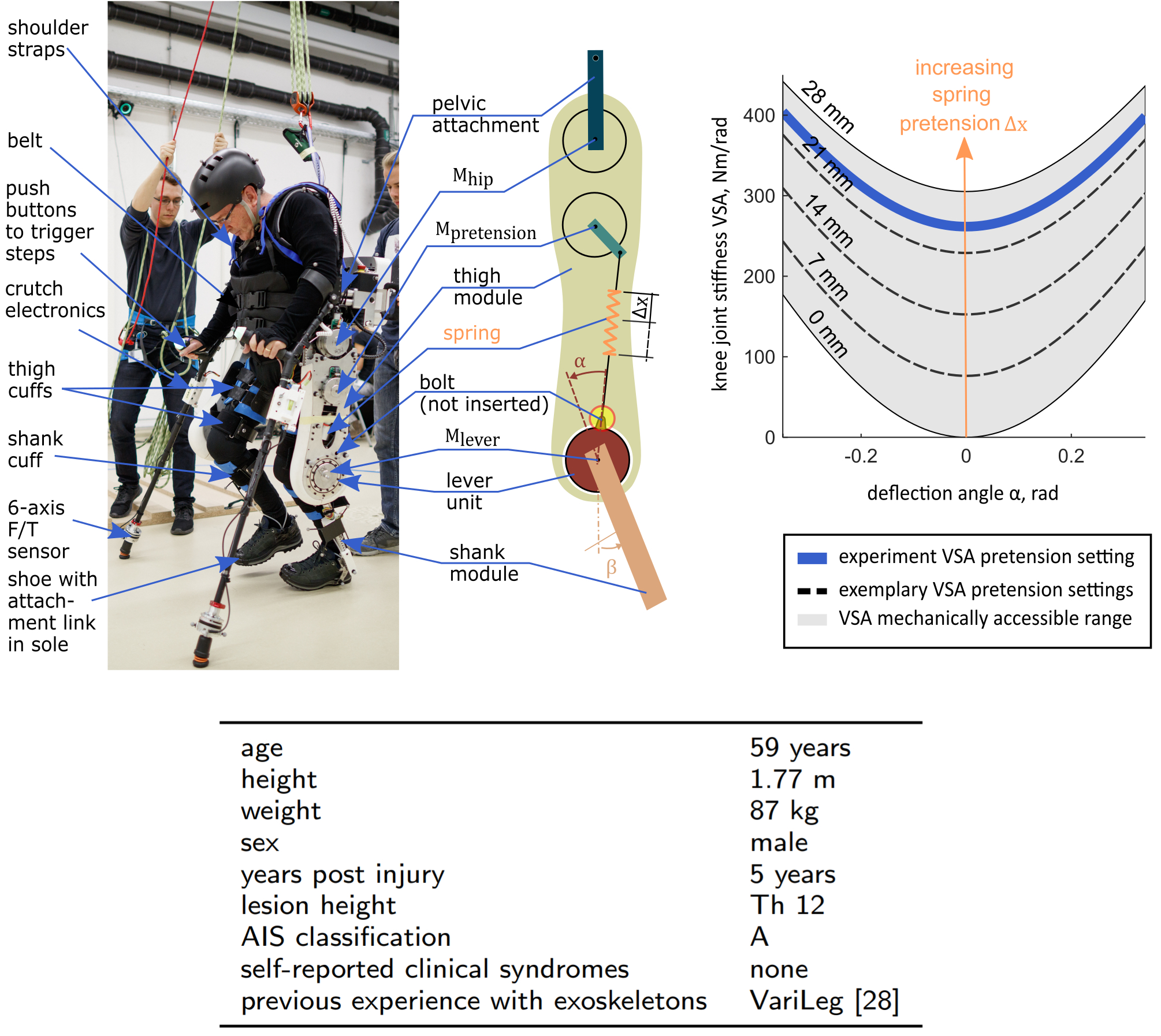}
    \caption{\textbf{Exoskeleton overview and case subject demographics} The exoskeleton has actuated hip and knee joints in the sagittal plane (left). The VSA in the knee joint can be blocked by inserting a bolt. If the VSA is active (bolt not inserted) $M_{pretension}$ can change the pretension of the spring $\Delta x$ to adjust rotational stiffness of the knee joint. The (constant) chosen stiffness of the VSA for the compliant configuration is indicated on the right side (blue line), and is in a similar range as the knee stiffness values during unimpaired walking~\cite{Pfeifer2014}. Demographics of the subject are listed at the bottom.}
    \label{fig:system_overview}
\end{figure*}

The VariLeg exoskeleton is a powered lower limb exoskeleton \cite{Schrade2018}. In short, it features actuated hip and knee joints allowing active flexion and extension, and a passive spring-loaded ankle joint to support passive dorsi- and plantar-flexion (Fig.~\ref{fig:system_overview}). The knee joints can be rigidly actuated (conventional actuation), or by a Variable Stiffness Actuator (VSA) inspired by the MACCEPA mechanism~\cite{VanHam2007} allowing online adaptation of knee joint stiffness. This allows for the testing of the two stiffness configurations on the exact same exoskeleton, thereby removing the potential bias that could come from comparing two separate devices with different designs, interfaces to the user or technical performances. To switch between the two configurations, a bolt is added to or removed from the system~(Fig.~\ref{fig:system_overview})~\cite{Schrade2019}.

The user is attached to the exoskeleton through custom-made shoes, cuffs at the shanks and thighs, a belt around the pelvis, and shoulder straps for the torso~\cite{Meyer2019}. The exoskeleton moves the legs according to pre-programmed trajectories that can be changed in terms of step length and step duration. Steps are triggered by the user with push buttons installed on the crutches. In the configuration where the VSA was used, the knee joint stiffness was set to a constant value of 260~Nm/rad (at 0$^{\circ}$ deflection) based on the thermal limitations of the motor. More details on the VariLeg design, control and performance evaluation can be found in previous work~\cite{Schrade2018}.

\subsection*{Subject Recruitment and Demographics}
This study was approved by the local ethics committees (BASEC Nr.~Req-.~2018-00568 and EK~2018-N-87). One subject (Fig.~\ref{fig:system_overview} bottom) was recruited and gave consent to participate after being informed about potential risks. The subject was experienced in walking with the VariLeg exoskeleton as he already participated in 52~training sessions of approximately one hour duration each in preparation for the CYBATHLON~2016~\cite{Schrade2018}. 

\subsection*{Experimental Protocol}
\begin{figure*}
    \centering
    \includegraphics[width=0.96 \textwidth]{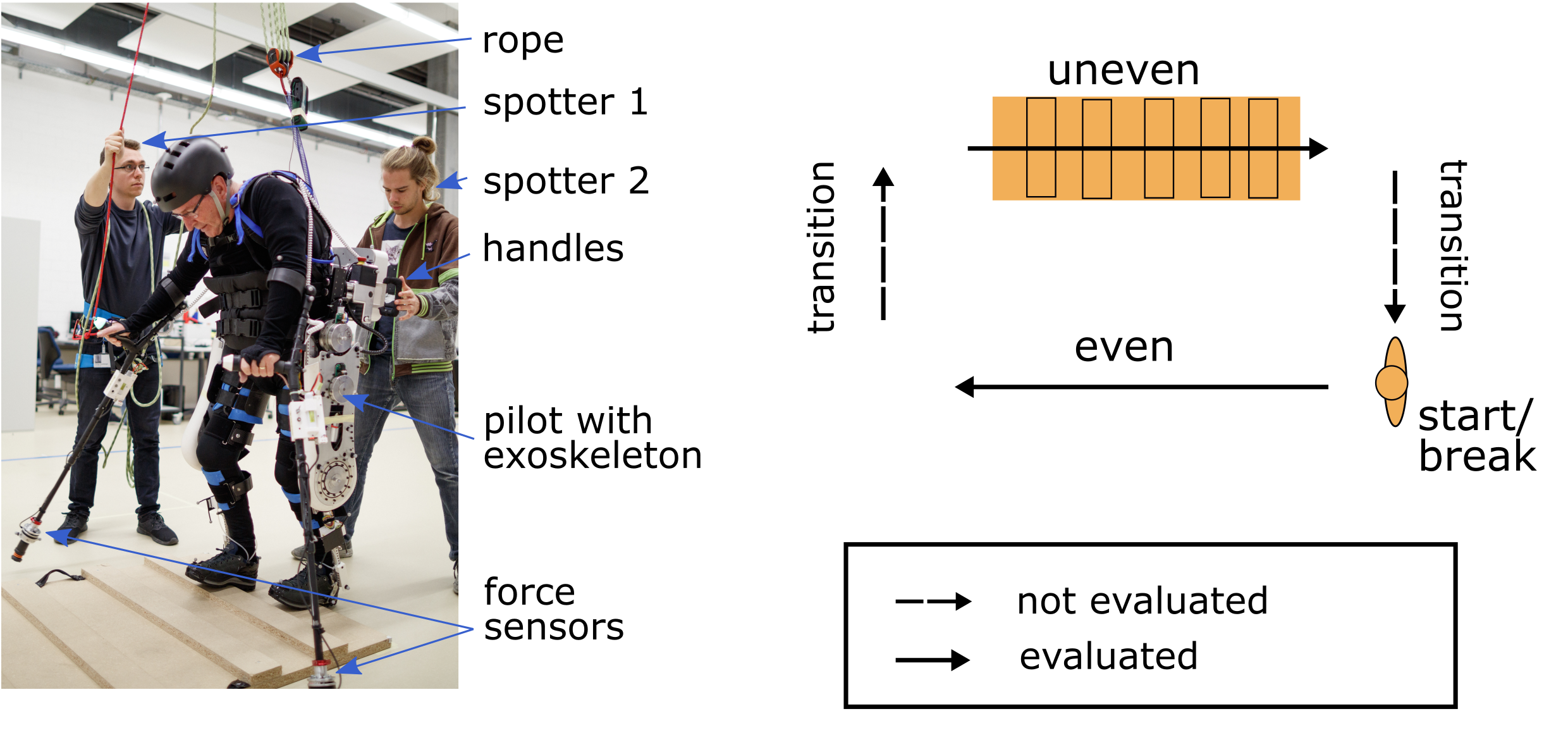}
    \caption{\textbf{Experiment Setup and Protocol Illustration} Spotter 1 operates the rope to prevent falling, spotter 2 can give support for minor corrections. Both spotters were instructed not to intervene during the measurements unless necessary (e.g., user losing balance) or if requested by the user. Every round of measurement consisted of one sequence of even and one sequence of uneven ground walking. The transitions between were not evaluated. Before every round, a break was administered.}
    \label{fig:experimental_setup}
\end{figure*}

This study was designed as a single case report with multi-session evaluation in a cross-over design and with extensive training over 10 sessions per condition. Rigid and compliant knee joints were compared on even and uneven ground walking with the same user piloting the exoskeleton. The even ground was a flat walkway of about 4~m length. Uneven ground was simulated by a wooden board of 2~m length with slats of about 0.03~m height placed at irregular distances as depicted in Fig.~\ref{fig:experimental_setup}.

Two spotters~(Fig.~\ref{fig:experimental_setup} left) ensured safety by walking behind the user. Spotter~1 used a harness to secure the pilot of the exoskeleton with an over-head fall prevention system. This ensured that the pilot could not fall if spotter~2 could not support all his weight via the handles in case the pilot lost balance. It is important to note that the over-head fall prevention system did not provide any weight compensation during walking with the exoskeleton. Additionally, spotter~2 only touched the handles at the pelvic structure (Fig.~\ref{fig:experimental_setup} left) if necessary during measurements. 

For each configuration (rigid and compliant), the user first received ten training sessions to familiarize with the exoskeleton configuration (familiarization phase). During these sessions, parameters influencing the execution of the trajectory were adapted to the user’s needs/requests to ensure optimal comfort and performance. The modified parameters included vertical distance between ankle and hip during stance (influencing knee flexion angle during stance phase) and torso inclination during double support (influencing rotation of ankle trajectory around reference point of the hip joint). Step length setting and step execution duration were kept constant. Following familiarization, measurements were taken on three consecutive sessions (measurement sessions) on different days. During one measurement session, ten sequences of even walking and ten sequences of uneven ground walking were collected. We recorded even and uneven ground in alternating fashion, always resting for a minimum of three minutes after collecting a sequence of each ground type~(Fig.~\ref{fig:experimental_setup} right). Additional breaks were performed at the user’s request. The compliant configuration was tested first (10+3 sessions), followed by the rigid configuration (10+3 sessions). Each session lasted for about 1 hour.

\subsection*{Data collection and analysis}
Optical tracking (Optitrack and Motive 2.0, NaturalPoint Inc., USA) was used to record the movement of the arms and crutches, torso, pelvis, thighs and shanks at 120~Hz. Forces and torques at both crutch tips were recorded with OMD-45-FE-1000N sensors (Optoforce, HUN) at 100~Hz. The custom-made instrumented crutches were connected with the exoskeleton via cable to record the force and torque data and for the pilot to trigger steps with push buttons~(Fig.~\ref{fig:system_overview}~left). Spotter intervention via the handles was monitored via touch sensitive sensors on the handles and engagement of the rope was observed with a force gauge installed in-line with the rope used to secure the user. This allowed excluding sequences from data analysis if spotters had interfered.

Motion capture data were labelled and processed using Motive (version 2.1, NaturalPoint Inc., USA). A  GCVSPL filter (quintic spline, unknown variability)~\cite{Woltring1986} was applied before analyzing joint angles and calculating transformation matrices in Visual3D (version 5, C-Motion Inc., USA). Data synchronicity between the exoskeleton log and the motion capture was ensured by recording a shared trigger signal on both data streams. Heel strikes were identified by calculating ankle marker vertical velocity, applying a threshold and correcting occurrence with ankle vertical position. Data were aligned from left heel strike to left heel strike. Subsequently, all strides were grouped according to ground and exoskeleton configuration, resulting in four groups: rigid on even ground (RE), rigid on uneven ground (RU), compliant on even ground (CE) and compliant on uneven ground (CU). 

For the analysis, parameters were calculated over strides or steps. Gait cycles (left strides) were divided into four temporal phases. The initial double support phase lasted from left heel strike to right toe off and was followed by single stance phase (with respect to the left leg). After right heel strike, which terminated the single stance phase, the final double support phase lasted until left toe off, which initiated swing phase (of the left leg). We extracted spatial parameters, namely stride length and step length, and temporal parameters from the recorded motion capture data. Average walking speed was calculated by measuring the distance the pelvis travelled during one gait cycle and dividing it by the cycle duration. We calculated trunk angle (lean), (whole-body) inclination angle, hip and knee angles as well as their respective ranges of motion (ROM). Foot trajectories based on the ankle marker position were also extracted, as well as the angle in the sagittal plane included between the two feet relative to the pelvis at heel strike (included angle, as depicted in~Fig.~\ref{fig:joint_included_angle}a). Kinematics of the pelvis in the transverse plane and the ratio of stride length to respective antero-posterior pelvis displacement (SLAP) were calculated.

From the measured crutch force data, we calculated the total net force (force applied to both crutches in all three directions) as well as its corresponding impulse (integral over time) over a gait cycle. We assumed the impulse to be a proxy measure of the user's effort to balance and ambulate~\cite{Carpentier2010}. We also derived the sum of left and right crutch forces along the medio-lateral, antero-posterior and vertical directions by rotating the crutch forces according to the crutch orientation in space identified by the motion capture system. Crutch forces of the individual crutches (left and right) are also reported in medio-lateral, antero-posterior and vertical directions together with their respective impulses.

We further analyzed where crutches were placed during left and right stance as well as the timing of crutch lift-off (crutch double support versus crutch single support). We calculated the base of support as the area of the triangle composed of the tips of the crutches and the midpoint of the stance foot as vertices.  Lastly, we calculated the Robinson symmetry index according to~\cite{Viteckova2018} for step length, included angle, and single and double support phase to quantify the symmetry between left and right steps.

Statistical analysis was performed with R (Version 3.6.0, R Foundation, Austria). Descriptive statistics and multiple linear regression models (Appendix \ref{LM}) were used to describe and estimate the effects of the two configurations and of the two ground types on the outcome parameters. The level for significance was set to p~$<$~0.05.

\section*{Results}
We analyzed 170 and 56 strides on even and uneven ground, respectively, when the exoskeleton was compliant. With rigid knee joints, we analyzed 232 and 101 strides for even and uneven ground respectively. Descriptive statistics and results from the linear regression models are summarized in Table~\ref{tab:descriptive_statistics} (Appendix \ref{LM}) and Table~\ref{tab:LM_results} (Appendix \ref{LM}) respectively.

\begin{figure*}[htbp]
    \centering
    \includegraphics[width=0.95 \textwidth]{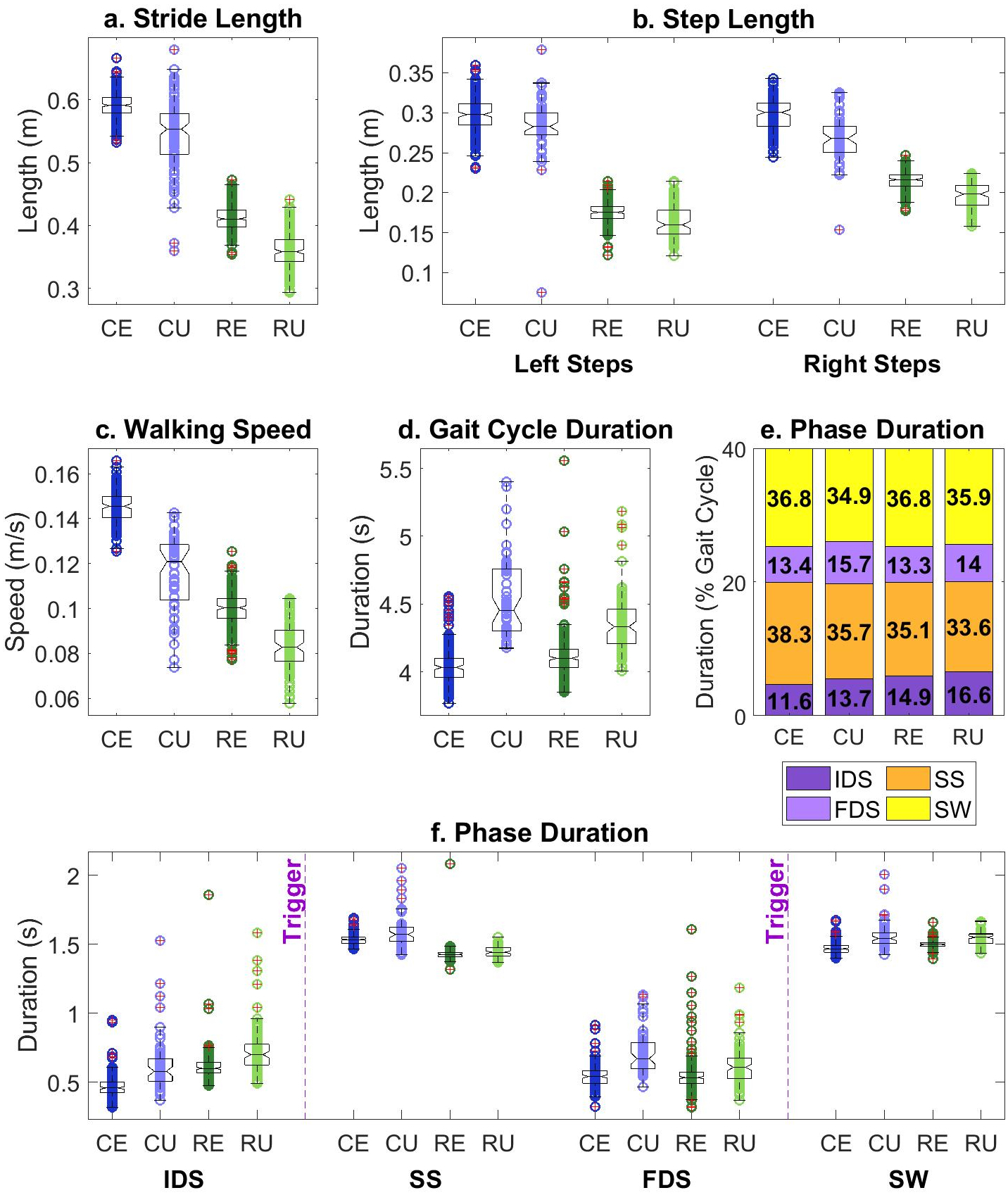}
    \caption{\textbf{Walking speed, spatial and temporal parameters for walking over two ground types with a rigid and compliant knee joint} Stride length decreased on uneven ground~(a)  as did walking speed~(c). Walking speeds were always higher for compliant knee joints, irrespective of ground type. Step length were generally shorter for the rigid configuration and on the uneven ground~(b) and gait cycle duration in seconds increased on uneven ground~(d). The gait cycle was segmented into initial double support~(IDS), single stance~(SS), final double support~(FDS) and swing phase~(SW). Steps were triggered by pushing a button during IDS and FDS phases~(f). The relative duration of the double support phases in gait cycle percent was higher on uneven ground~(e).}
    \label{fig:spatial_temporal_parameters}
\end{figure*}

The compliant configuration enabled higher walking speeds ($\mu_{CU}$: 0.116~m/s, $\mu_{CE}$: 0.145~m/s) than the rigid configuration ($\mu_{RU}$:~0.083~m/s, $\mu_{RE}$:~0.100~m/s) (p~$<$~0.001). Generally, walking speed always decreased when walking over uneven ground~(Fig.~\ref{fig:spatial_temporal_parameters}c). Walking speed was directly influenced by stride length (Fig.~\ref{fig:spatial_temporal_parameters}a). Right and left step length (Fig.~\ref{fig:spatial_temporal_parameters}b) were influenced by the exoskeleton configuration~(p~$<$~0.001) and by ground type~(p~$<$~0.001).  Gait cycle duration (Fig.~\ref{fig:spatial_temporal_parameters}d) increased on uneven ground compared to even ground~(p~$<$~0.001). Configuration slightly influenced gait cycle duration~(p~$<$~0.001). Cycles were slightly longer on uneven ground with the compliant configuration compared to the rigid~(p~$<$~0.001), whereas the opposite was the case for even ground walking~(p~$<$~0.001).

The increase in gait cycle duration by 0.18 seconds on average on the uneven ground was mostly caused by longer initial and final double support phases. Single stance and swing phase were slightly influenced by ground type~(p~$<$~0.05, increasing duration on uneven) and configuration~(p~$<$~0.01, decreasing single stance and increasing swing phase with rigid knee joints) (Fig.~\ref{fig:spatial_temporal_parameters}f).

\begin{figure*}[htbp]
    \centering
    \includegraphics[width= 0.80 \textwidth]{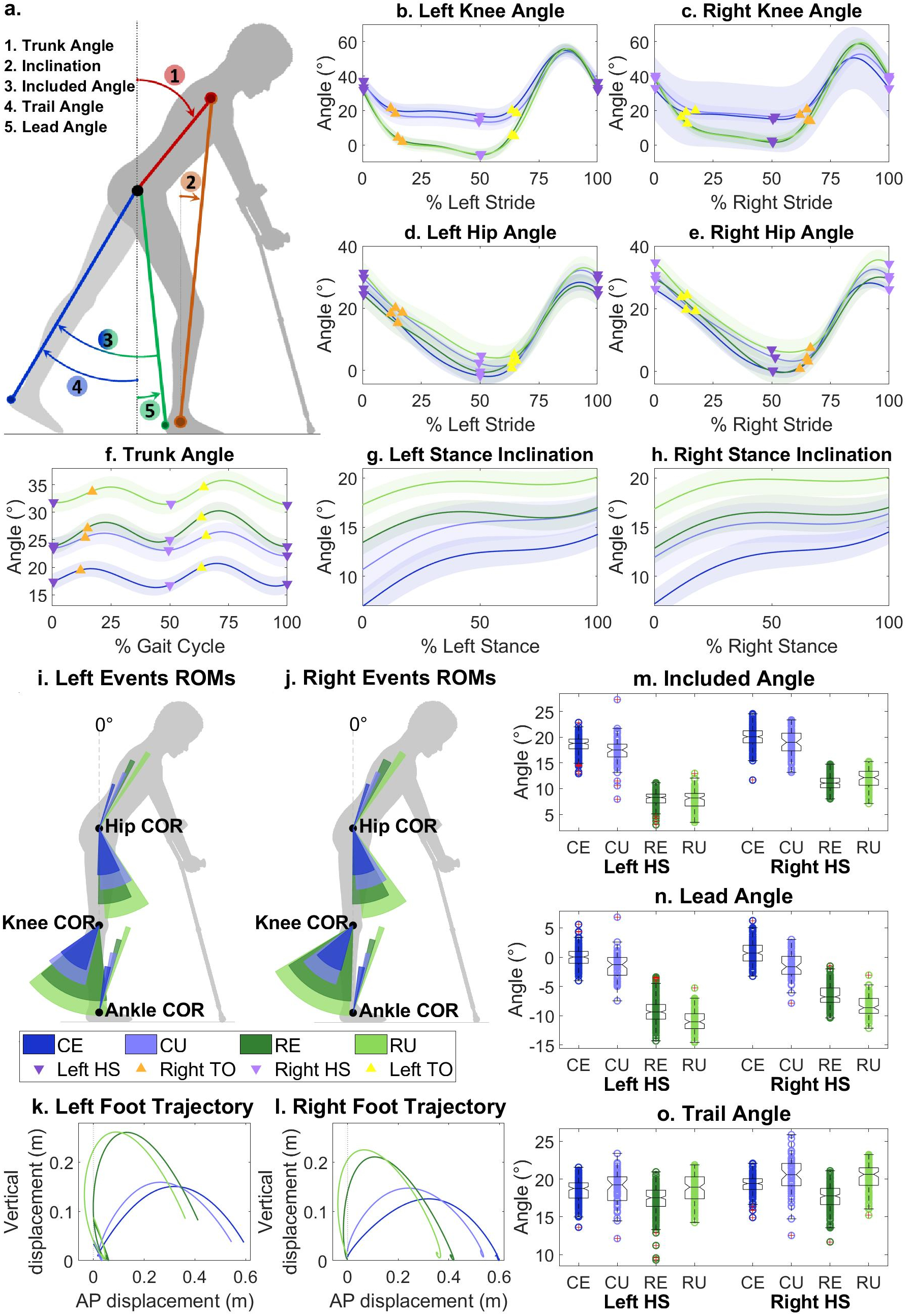}
    \caption{\textbf{Illustration of joint angles, ranges of motion and foot trajectories during one gait cycle} A schematic of the analyzed angles is provided in (a). Average trajectories of knee angle (b-c), hip angle (d-e), trunk angle~(f) and inclination~(g-h) are indicated with lines plus minus standard deviations as shaded areas. Triangles represent mean heel strike (HS) and toe off (TO). Joint ROMs are displayed for left (i) and right (j) events (steps, stances, strides). The inclination ROM has the ankle as center of rotation (COR). Hip ROM varied with ground type but not with configuration. Knee ROM varied with configuration but not with ground type. More knee extension could be observed with a rigid exoskeleton. Feet trajectories relative to the lab space are displayed in (k) and (l).  
    Included angle~(m), lead angle~(n) and trail angle~(o) at heel strike (HS) are displayed in boxplots. Whiskers indicate the minimum and maximum sample not considered outliers. Outliers are marked with crosses. Box edges indicate the 25th and 75th percentile, the center line represents the median and the notches extend to the 95\% confidence interval of true differences between means. Included angle was higher with compliant knee joints but not influenced by ground type. Lead angle varied with ground type and knee configuration whereas trail angle seemed to be rather constant.}
    \label{fig:joint_included_angle}
\end{figure*}

The included angle was smaller for the rigid configuration~(p~$<$~0.001) (Fig.~\ref{fig:joint_included_angle}m). The difference was mainly caused by a decrease of the lead angle (Fig.~\ref{fig:joint_included_angle}n), as the trail angle (Fig.~\ref{fig:joint_included_angle}o) stayed rather constant across configurations. The trail angle was slightly higher on uneven ground.

The inclination angle was increased on uneven ground compared to even ground~(Fig.~\ref{fig:joint_included_angle}g-h)~(p~$<$~0.001). Rigid knee joints also caused an increase in inclination angle compared to compliant ones ($\Delta_{peak\ angle}$~$>$~2.6$^{\circ}$, p~$<$~0.001). The inclination angle for the compliant configuration on uneven ground ($\mu_{peak\ angle}$ of 16.9$^{\circ}$ and 16.6$^{\circ}$ for left and right stance respectively) was almost the same as the inclination angle for the rigid configuration on even ground~($\mu_{peak\ angle}$~of 17.1$^{\circ}$ and 17.2$^{\circ}$ for left and right stance, respectively). Trunk angle~(Fig.~\ref{fig:joint_included_angle}f) seemed to be affected by ground and configuration in a similar way to the inclination angle~(p~$<$~0.001).
The ROM of the knee joint~(Fig.~\ref{fig:joint_included_angle}i-j) was significantly larger in the case of a rigid exoskeleton joint compared to a compliant one~($\Delta$~$>$~14.623$^{\circ}$, p~$<$~0.001). It did not considerably vary with ground type, however. The left and right knees were asymmetric in the range of angles that were spanned for both exoskeleton configurations and ground types. The rigid right knee was extending further than the left knee. For the right knee, variability was rather large on uneven ground. This was not the case for the left knee on uneven ground when it was compliant~(Fig.~\ref{fig:joint_included_angle}b-c). The foot trajectory in the sagittal plane~(Fig.~\ref{fig:joint_included_angle}k-l) indicated that the rigid configuration resulted in more ground clearance but shorter steps.

\begin{figure*}
    \centering
    \includegraphics[width=0.94 \textwidth]{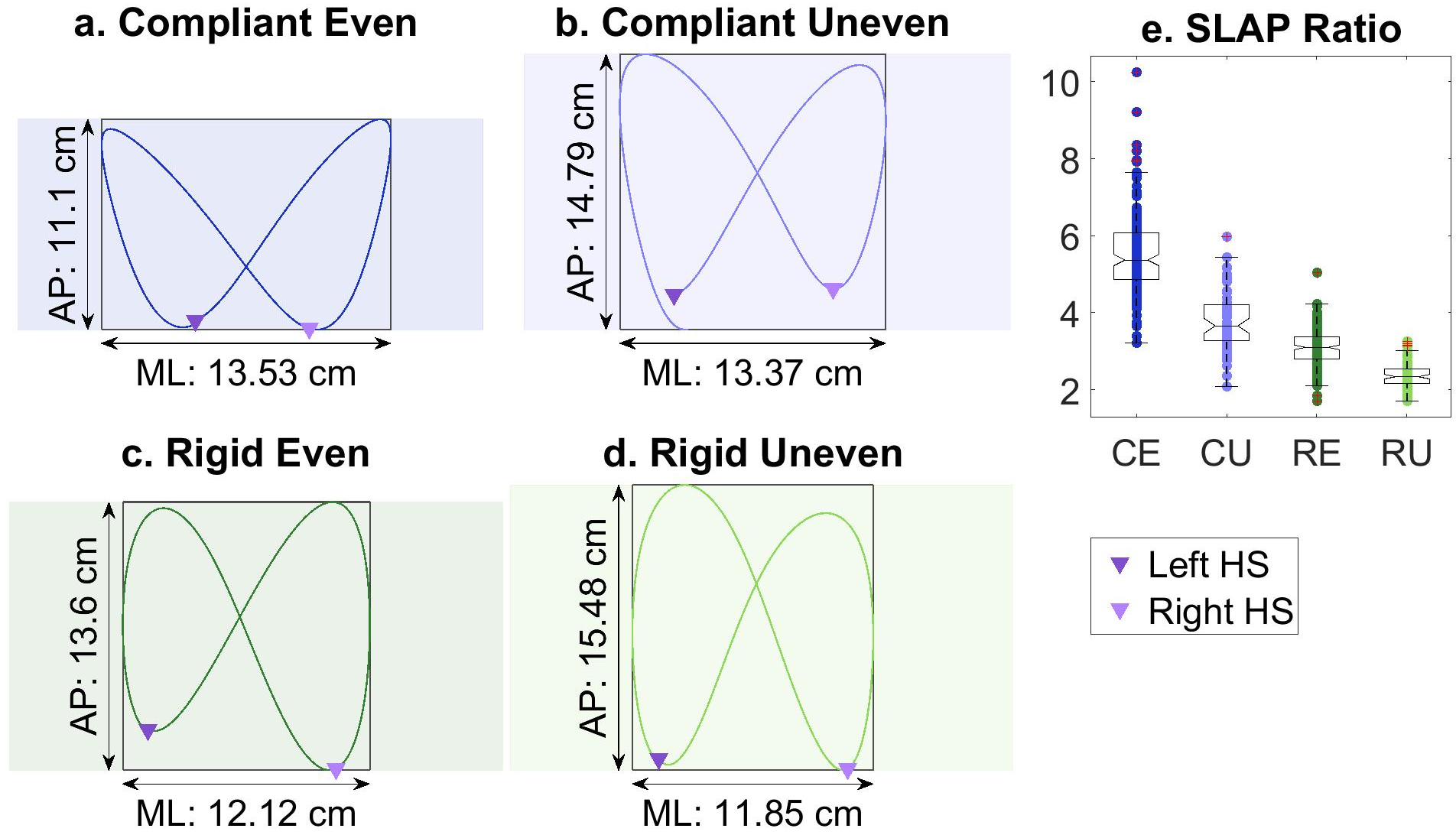}
    \caption{\textbf{Pelvis movement in the transverse plane with respect to step width} Shaded box width indicates step width and the trajectories illustrate movement of the pelvis (a-e). Heel strike (HS) of left and right foot are indicated with triangles. Medio-lateral (ML) movement was bigger with a compliant exoskeleton. Anterior-posterior (AP) movement was smallest with a compliant exoskeleton on even ground. SLAP ratios are shown in (e). Stride length was biggest in relation to the AP movement with compliant knees on even ground. For the other combinations it gradually decreased.}
    \label{fig:Trunk_CoM}
\end{figure*}

The pelvis movement in the transverse plane was increased in the anterior-posterior (AP) direction when walking over uneven ground~(p~$<$~0.001, Fig.~\ref{fig:Trunk_CoM}).
Stride length compared to AP movement (SLAP ratio, Fig.~\ref{fig:Trunk_CoM}e) was largest with a compliant exoskeleton on even ground~($\mu_{CE}$ = 5.53). When walking over uneven ground, stride length decreased in relation to the AP movement. It was even further reduced with a rigid exoskeleton on even ground to the smallest value with rigid joints on uneven ground~($\mu_{RU}$ = 2.37).
Medio-lateral (ML) movement slightly decreased when walking over uneven ground for the compliant exoskeleton (p > 0.05). It further decreased when the exoskeleton was rigid~(p~$<$~0.001). Walking on uneven ground with a rigid exoskeleton resulted in the smallest ML movement.

%Crutch Placement

\begin{figure*}
    \centering
    \includegraphics[width=0.97 \textwidth]{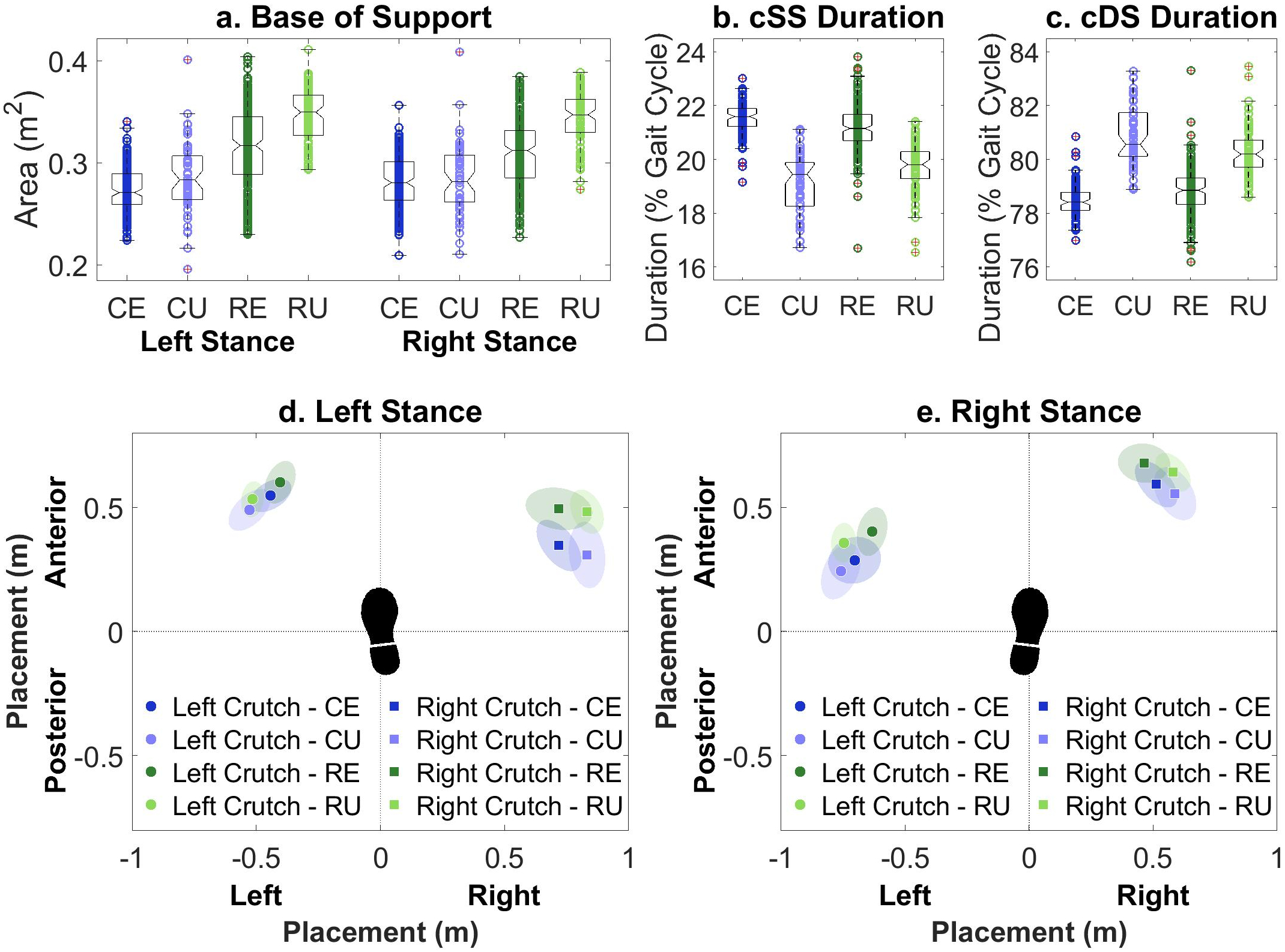}
    \caption{\textbf{Crutch placement and base of support} The base of support (a) was calculated as the area of the triangle between both crutches and the stance foot during stance. It most notably increased when switching from compliant knee joints to rigid ones. The base of support significantly increased on uneven ground for both configurations. The percentages of a gait cycle where only one crutch was in contact with the floor (cSS) and where both crutches were on the ground (cDS) are given in (b) and (c), respectively. The user had both crutches on the ground for a longer portion of the gait cycle when on uneven ground. The averaged positioning of the crutches with respect to the stance foot is shown in (d-e). Shaded ellipses represent the 95\% confidence interval of data points.}
    \label{fig:Crutch_Placement}
\end{figure*}

The crutches were placed more anteriorly with a rigid exoskeleton (Fig.~\ref{fig:Crutch_Placement}d-e). Uneven ground mainly led to more lateral placement, resulting in a slightly more posterior placement than on even ground.
On uneven ground, the base of support (Fig.~\ref{fig:Crutch_Placement}a) did not significantly increase for the compliant exoskeleton right stance (p = 0.8) but it significantly increased for left stance~(p~=~0.017). With a rigid exoskeleton  the base of support was increased on even ground compared to the compliant configuration~(p~$<$~0.001), and further increased when walking over uneven ground~(p~$<$~0.001).

The gait cycle percentage during which one of the crutches was in the air for repositioning (crutch single stance (cSS) duration) decreased on uneven ground~(p~$<$~0.001) (Fig.~\ref{fig:Crutch_Placement}b). It was lowest on uneven ground with compliant knee joints~(18.93\%) and longest for compliant knee joints on even ground~(21.53\%). The part of the gait cycle during which both crutches were on the ground (crutch double support (cDS) duration) increased for uneven ground accordingly~(p~$<$~0.001) (Fig.~\ref{fig:Crutch_Placement}c).

\begin{figure*}[htbp]
    \centering
    \includegraphics[width=0.8 \textwidth]{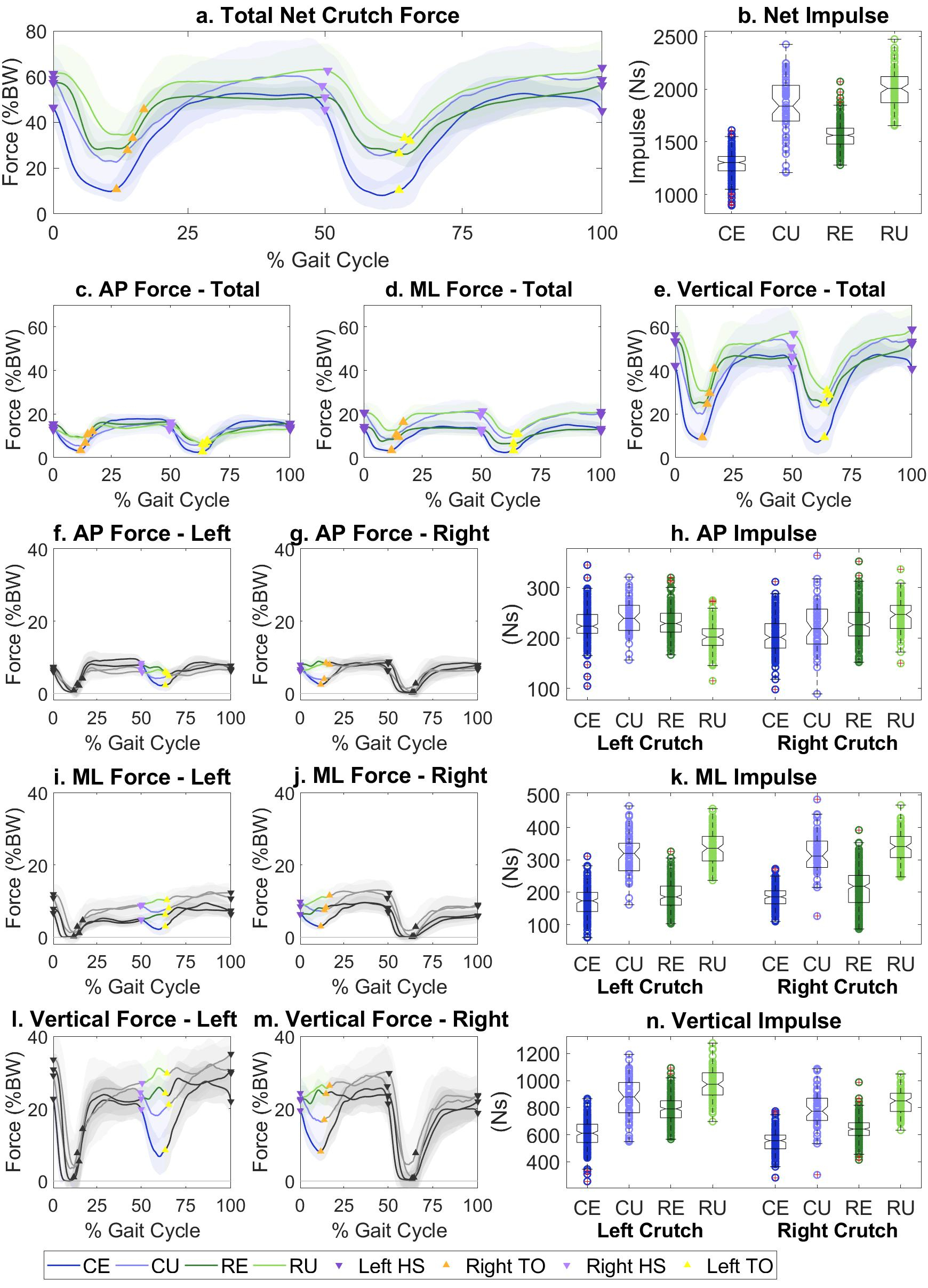}
    \caption{\textbf{Crutch forces (normalized with respect to body weight) and impulses} The force applied to both crutches in all three directions, named total net crutch force, is shown in (a). Total net crutch force was higher on the uneven ground. The compliant configuration resulted in lower forces during the crutch unloading phases for both ground types. Impulse from the total net force (b) was lowest for compliant knees and even ground. Walking over uneven ground significantly increased impulse for both configurations. The components of the total net force in the anterior-posterior  (c), medio-lateral (ML) (d) and vertical (e) directions are also displayed. 
    The forces applied in the AP, ML and vertical direction to the left (f, i, l) and right (g, j, m) crutch are reported too. The corresponding impulses (h, k, n) were also calculated. With a compliant exoskeleton, during unloading and repositioning of one crutch the contralateral crutch was unloaded, which did not occur with rigid knees. This timespan is represented by the colored parts of the force averages. The impulse of the vertical force was lower for the right crutch compared to the left side for both configurations and both ground types.
    Heel strike (HS) and toe off (TO) are indicated with triangles. }
    \label{fig:all_forces}
\end{figure*}

Total net crutch force showed higher maximum amplitudes for both configurations on the uneven ground ($\Delta$~$>$~135 N, p~$<$~0.001) (Fig.~\ref{fig:all_forces}a). The difference between configurations on even ground is still statistically significant but smaller ($\Delta$~$=$~15 N, p~$<$~0.01). After heel strike, the user unloaded the crutches. Shortly before the next step was executed, the crutches were loaded again. With the compliant knee joints, lower forces were observed during the initial and final double support phases compared to rigid ones on the same ground. The impulse (Fig.~\ref{fig:all_forces}b) increased from even to uneven ground, and compliant to rigid exoskeleton, respectively~(p~$<$~0.001). The most notable differences between the configurations occurred in the total vertical crutch force component (Fig.~\ref{fig:all_forces}e). These forces indicated that on uneven ground 50-60\% body weight were unloaded onto the crutches for both configurations. Total ML forces (Fig.~\ref{fig:all_forces}d) showed the same tendency but were of lower amplitude than the vertical forces. Total AP force (Fig.~\ref{fig:all_forces}c) was very similar for both ground types within the same exoskeleton configuration.

The results were similar when the individual crutches were analyzed. The most notable difference between the configurations was observed in the vertical crutch force component (Fig.~\ref{fig:all_forces}l-m). During repositioning of one crutch, with a compliant exoskeleton the contralateral crutch was unloaded. The vertical impulse (Fig.~\ref{fig:all_forces}n) was lower with a compliant exoskeleton on the respective ground type~($\Delta_{CE-RE}$~$>$~95.8~Ns and $\Delta_{CU-RU}$~$>$~54.6~Ns, p~$<$~0.001). The vertical impulse of the right crutch force was always lower than the one of the left crutch force irrespective of configuration or ground type.

The symmetry indices were generally nearer to zero, which indicates perfect symmetry, if the exoskeleton was compliant~(Table~\ref{tab:symmetry_indices}). Especially, step length and included angle showed high symmetry indices for walking on even ground with a rigid exoskeleton, which indicates high asymmetry.

\begin{table}[hbt]
	\caption{Symmetry indices of selected parameters}
	\label{tab:symmetry_indices}
	\centering
	\begin{tabular}{l r r r r}
	
		\toprule
                        & CE       &   CU     &   RE     &   RU     \\
                        		\midrule
step length             & 0.1      &   5.3     &   20.6    & 18.6    \\
included angle          & 7.8    & 9.5    & 31.99    & 41.1     \\
single support phase    &   4.0    &   2.1    &   4.8   &   6.6   \\
double support phase    &   14.4      &   14.5  &   11.9    &   16.6   \\
		
		\bottomrule
	\end{tabular}
\end{table}

\section*{Discussion}

We aimed to investigate how knee joint compliance influences performance, efficiency and effort of a user with motor complete SCI while walking over even and uneven ground with a powered lower limb exoskeleton.
We found that walking speed was slower with rigid joints, despite presenting larger knee range of motion. We argue that this is an indicator for decreased  efficiency since larger limb excursions require more energy and should be avoided if that movement is not translated into forward motion during ambulation.
Decreased loading of the crutches could not be observed as clearly as expected in terms of measured peak crutch force amplitudes. However, the impulse of the crutch forces over one gait cycle decreased significantly for compliant knees compared to conventional rigid actuation. From this we infer that it took more effort to walk with a rigid exoskeleton, as more force was needed over a gait cycle. This is based on the assumption that net force created by the user correlates with effort spent by muscles of the upper body and therefore has an influence on metabolic cost. Additionally, we also observed greater asymmetry between left and right steps when the knee joints were  actuated rigidly.

On uneven ground, walking speed was generally lower compared to even ground walking. This was expected because of the higher difficulty to maintain balance and establish ground clearance of the swing leg irrespective of knee joint configuration. It is noteworthy that walking speed with rigid knee joints on even ground was even slower than walking speed over uneven ground in the compliant configuration. We observed that the impulse of the total net crutch force increased for walking over uneven ground, which indicates more effort was necessary to traverse it compared to the even ground.

\subsection*{Spatial and temporal parameters}
The mean walking speed over all sessions and trials measured for the conditions during this study ranged from 0.083~m/s (rigid on uneven ground) to 0.145~m/s (compliant on even ground), placing it in the lower range of the reported mean speed (0.06~m/s to 0.27~m/s) reached by non-ambulatory SCI subjects walking with the EksoGT on even ground~\cite{Ramanujam2018b}. Typically achieved walking speed highly depends on the capabilities of the used exoskeleton, and on the lesion level and physical condition of the user. In another study, users were reported to achieve similar stride lengths to the longest strides of around 0.6~m observed in this study, resulting in a slightly faster walking speed of around 0.2~m/s using the EksoGT~\cite{Ramanujam2018}. However, these users all suffered incomplete SCI allowing them to partially assist the exoskeleton movements, which may explain the higher walking speed achieved. Interestingly, able-bodied subjects were also evaluated as a reference when walking with the EksoGT, achieving walking speeds from 0.25~m/s when the users were instructed to be  passive and 0.36~m/s when they actively supported step execution with their legs. Consequently, walking speed still is a limitation of exoskeletons since motor-complete users achieve gait speeds of only 0.26~m/s~\cite{Louie2015} on average. This is lower than the speed of 0.4~m/s, which is considered the minimum for limited community ambulation~\cite{VanHedel2008}. Consequently, any design choice that can increase speed should be welcomed.

Walking speed is determined by two parameters, namely stride length and gait cycle duration. Ramanujam et al.~\cite{Ramanujam2018} found that the speed reached by SCI subjects walking in the EksoGT strongly positively correlated with stride length. This was also observed in our study. The compliant knees allowed for longer strides and resulted in higher walking speed. Figure \ref{fig:spatial_temporal_parameters}.a-c shows that mean walking speed and mean stride length follow the same trend across conditions. Mean stride length and step length measured with compliant knees (0.55-0.60 m and 0.27-0.30 m respectively) were similar to the ones achieved on even ground in~\cite{Ramanujam2018} (0.59-0.64 m and 0.27-0.32 m), while the ones observed with rigid knees were lower (0.36-0.41 m and 0.16-0.22 m). The observed decrease in stride length of the rigid exoskeleton in comparison to the compliant one might have been influenced by two factors: altered system behavior despite identical control, and the user's walking strategy. Since the compliant joint always allows deviations from the desired position, we expected that knee flexion would be reduced due to gravity during swing, which would reduce step length as heel strike would occur earlier. In contrast, we would have expected that the increased position control fidelity of the rigid configuration would lead to more accurate tracking of the reference trajectory and thus would have resulted in longer steps with a rigid knee joint. However, we observed the opposite in our results and steps were longer in the compliant configuration. Talaty et al.~\cite{Talaty2013} compared different SCI users walking with the ReWalk and observed that subjects who could adopt a body posture that allowed enough foot clearance performed longer steps. In our experiment, however, foot clearance did not appear to be the limiting factor, because it was higher for rigid knee joints even though the resulting steps were shorter. Even more, since the feet moved more in the vertical direction without progressing in the anterior direction, the higher foot clearance was ineffective and therefore also inefficient. The observed higher foot clearance could have been influenced by the faster response of the knee motor to desired position changes, in absence of a delaying spring. The user's walking strategy adopted with the rigid configuration led to reduced lead angle (decreasing step length) as he increased his inclination angle. The included angle between lead and trail leg was thus decreased as the trail angle did not change significantly, which is assumed to in turn have reduced the step length. Other studies reported that the included angle positively correlated with stride length~\cite{Ramanujam2019}. Comparing the mean lead and trail angles obtained in this experiment (-10.75-0.77$^\circ$ and 17.30-20.53$^\circ$, respectively) with the previously mentioned study (around 3-9$^\circ$ and 16-24$^\circ$), we can see that the trail angles were similar, while the lead angles were noticeably smaller in our case. Most likely, the fact that the pilot was leaning further forward prevented the lead foot from being placed further in front of the body. This was most prominent with rigid knee joints and resulted in negative lead angles and, consequently, in smaller included angles and shorter strides.

The other parameter influencing walking speed is gait cycle duration. In this study, the theoretical swing time planned by the exoskeleton was set to 1.75~s, and therefore the speed at which the exoskeleton moved the feet of the pilot along the set trajectory to perform a step was always the same. Despite this, the subject could walk faster with compliant knee joints than with rigid ones on both ground types. One factor influencing gait cycle duration could have been that the subject could actively influence the duration of the double support phases, since he had to manually trigger the next step. Ramanujam et al.~\cite{Ramanujam2018} suggested that the ability of the pilot to shift the weight from one foot to the other during the double support phase strongly influences the walking speed. Therefore, the higher speed obtained with compliant knee joints might suggest that, in this configuration, the subject could transfer the weight in a more efficient way, shortening the double support phase. However, according to the results in Fig.~\ref{fig:spatial_temporal_parameters}f, this happened only during initial double support phases, but was not the case for final double support phases. Compliant knee joints even resulted in the longest final double support phases on uneven ground despite allowing higher average walking speed than rigid knee joints on even and uneven ground. The shortest time spent in double support was observed for compliant knee joints on even ground (25\% of the gait cycle).
In contrast, the longest time was spent in double support on uneven ground with rigid knee joints (30.6\% of the gait cycle). The duration of the double support phases were similar or shorter than double support phases reported for incomplete SCI subjects walking with the ReWalk (30.49\% gait cycle) and with the EksoGT (58.89\% gait cycle) after training~\cite{Ramanujam2017}. This difference is noteworthy as our measurements were performed with a motor-complete SCI user, which typically walk slower than incomplete SCI users. This parameter could, however, also be influenced by the method used to define heel strike and toe off time, which proved to be non-trivial for exoskeleton walking, and the methods used in other work (e.g.~\cite{Ramanujam2017}) were not documented.

\subsection*{Joint angles and joint range of motion during walking}
For all four conditions, the mean trunk peak angle~(19.80$^\circ$–35.86$^\circ$) was higher than the one reported
in~\cite{Ramanujam2019}, which was below 10$^\circ$ for all the incomplete SCI subjects walking with the ReWalk or with the EksoGT. In our experiment, rigid knee joints led to higher trunk and inclination peak angles. This might be a strategy adopted by the user in order to avoid falling backwards and facilitate weight shifting. During the first training sessions with rigid knee joints, our participant attempted to walk more upright, but he could often not shift his weight over the stance foot to progress during the steps and therefore frequently lost balance.  This might be due to the greater vertical excursion needed to shift the center of mass over a rigid leg compared to a compliant one, where the knee can be flexed by loading the leg. Because of this issue, the pilot possibly opted for leaning forward with the whole body, thus bringing the center of mass more to the front of the feet and decreasing the probability of falling backwards. This behavior was accentuated on uneven ground, in both configurations. In this case, higher inclination and trunk angles could also result from the fact that the subject felt less safe. He might therefore have required more visual cues from the ground or opted to rely more on the crutches and less on the exoskeleton to better control balance. In all these cases, higher inclination and trunk angles caused the pilot to carry more weight on the crutches and less on the feet, therefore leading to higher effort for the upper body. This fact is confirmed by the higher impulses measured with rigid knee joints and on uneven ground (Figure~\ref{fig:all_forces}b).
The knee ROM was significantly higher on both ground types with rigid knee joints, mainly because during stance the knee was more extended (Figure~\ref{fig:joint_included_angle}b-c) compared to the compliant configuration, which showed increased knee flexion when loaded with the user's body weight. Consequently, the greater extension of the stance leg reached with rigid knee joints led to higher foot clearance for the contralateral foot (Figure~\ref{fig:joint_included_angle}k-l). Despite this, the steps were shorter, suggesting that the higher foot clearance in comparison to the compliant configuration was not useful for improving the step length, and was not necessary. As with compliant knee joints, the pilot could perform longer steps even if the knee ROM was smaller and the foot clearance lower. Hence, we could conclude that the compliant configuration was more efficient in generating forward movement since less leg excursion was necessary. 

\subsection*{Pelvis trajectory in the transverse plane}
The only other study investigating walking over uneven ground with exoskeletons~\cite{Ugurlu2014} found that trunk pitch angle velocity was increased if the exoskeleton was rigid. They suggested that when walking in a rigid exoskeleton, external forces, e.g. resulting from heel strike, have a more direct impact on the exoskeleton and, consequently, on the user. Thus, external forces may require the pilot to perform undesired forward-backward movements with the upper body to maintain balance. This may explain why during this experiment the AP displacement was higher with rigid knee joints on the same ground type, which did not allow the exoskeleton to weaken the impact that external forces exerted on the pilot, while compliant knee joints allowed this via the springs. The fact that the higher AP excursion was unnecessary is supported by the results of the SLAP ratio: lower SLAP ratios indicate that for the same AP displacement, the stride length was lower, therefore the AP excursion was not efficiently used to walk forward. Compliant knee joints led to higher SLAP ratios on both even and uneven ground, implying that higher stride length could be reached with lower AP excursion. We interpret this as being more efficient because the upper body had to be moved less in the AP direction per distance travelled. Since movement of the pelvis is assumed to be associated with effort of the user and becomes more difficult to control with higher lesion levels, we conclude that this ratio should be maximized for efficient exoskeleton walking. Overall, the results of this study are in agreement with the results of~\cite{Ugurlu2014}, and we also found that compliance helped walking over uneven ground. We could further demonstrate that this was also the case for a user with motor-complete SCI in overground walking and with crutches instead of a grounded bar to hold on to.

For the rather static walking pattern employed by most powered exoskeletons, shifting the body weight from one leg to the other to unload the swing leg is a skill that users need to be trained in specifically. Only sufficient weight shift clears the swing leg, ensuring successful swing through. Properly loading the stance leg also helps to unload the crutches, which have to bear all the weight that could not be shifted onto the stance leg. Healthy subjects walking in an exoskeleton have been reported to exhibit larger lateral weight shift in comparison to SCI users~\cite{Ramanujam2019} when using the same device. Thus, we assumed that the increased lateral movement observed with the compliant exoskeleton suggests that the user achieved lateral weight shift more easily, and the decreased lateral movement with a rigid exoskeleton pointed towards reduced ability to shift the weight onto the stance leg. Lastly, compliant knee joints resulted in a  more symmetrical and regular pelvis trajectory, similar to what is observed in able-bodied subjects walking with an exoskeleton~\cite{Ramanujam2019}.

\subsection*{Crutch-ground interaction forces and impulse}
On the same ground type, the net total  crutch force showed lower peaks with compliant knee joints. However, on uneven ground the difference was not statistically significant. When comparing the same ground type, the impulse resulting from the net total support ground reaction forces (GRF) was higher for the rigid configuration. Higher impulse means that, over time, the arms had to exert more force to maintain balance, potentially increasing user fatigue. This is especially relevant for users with higher lesion levels, who may not be able to fully support their trunk.

The lower impulse and GRF obtained in the compliant configuration are in line with the results in \cite{Ugurlu2014}, which showed that when walking with a compliant controller the force exerted by the arms was lower than when walking with a conventional one. 
Looking at the colored parts of the force curves in Figure~\ref{fig:all_forces}f-m, it can be observed that the biggest difference across conditions resulted from the double support phases of the gait cycle, when one of the two crutches was lifted to be repositioned. During these phases, the GRF decreased with compliant knee joints, while a plateau was observed in the rigid configuration. This trend is most prominent in the vertical force component. If we consider, for example, the final double support phase of the gait cycle (between right heel strike and left toe off), we can see that the right crutch was lifted to be placed forward during this phase, and the forces therefore approached~0~N. Interestingly, with compliant knee joints also the left (contralateral) crutch forces decreased. This means that the left arm could relax, while with rigid knee joints this was not observed, and the left arm had to continuously exert forces through the crutch. The same pattern was observed for the right crutch during initial double support, when the left crutch was lifted. This could imply that during the double support phases with compliant knee joints, the pilot relied more on the exoskeleton and less on the crutches when transferring the weight from one foot to the other. 
  
The impulse of the ML force component of the individual left and right crutch reflected the strategy used by the subject when walking, being higher for uneven ground, as the crutches were placed more laterally in this case. We expected the AP impulse to be higher for rigid knee joints, since the crutches were placed more anteriorly, but this was not the case. A reason for this could be that faster walking may increase the AP impulse of the crutches, as they are mainly used to brake the subject’s forward falling motion along this direction. However, the impulse was not corrected for walking speed, and the fact that with compliant knee joints the pilot walked faster probably contributed to increasing the AP impulse.

Minimizing impulse and GRF during walking with the exoskeleton is important because SCI subjects with high lesions potentially have less strength in the upper body. Exerting lower forces could prevent fatigue, and allow a longer walking time. Furthermore, impulse and GRF are transmitted to the shoulder joints and could potentially lead to overuse injuries. Consequently, reduced peak forces and impulse achieved with compliant knee joints are  an advantage in terms of both effort and joint injury prevention.

\subsection*{Crutch placement strategy}
Since the VariLeg only actuates the legs in the sagittal plane and does not have degrees of freedom in the frontal plane, neither pilot nor exoskeleton could increase the base of support to actively control lateral stability by performing wider steps as unimpaired walkers might do~\cite{Bauby2000}. Placing the crutches further away from the stance foot was the only possibility to increase the base of support for the user. Different studies in unimpaired subjects showed that step width, and therefore the base of support, was greater when balance was more challenging, as is the case on irregular surfaces~\cite{Marigold2008} or when closing the eyes~\cite{Bauby2000}. During our experiment, the base of support was smaller with compliant knee joints compared to rigid ones on both ground types. This could indicate that compliant knee joints allowed better handling of ground irregularities and required less engagement and therefore effort of the user's upper body to keep balance. Larger base of support with a rigid knee joint came from a more anterior placement of the crutches with respect to the stance foot, probably also related to the higher inclination angle observed. On uneven ground, the crutches were placed more laterally compared to even ground, which could indicate that higher lateral stability was needed. The placement strategy indicates that the crutches were mostly used to restrain lateral and forward motion and not to push the subject forward, therefore acting to prevent lateral and forward falling. This strategy has previously  been observed when subjects with incomplete SCI walk with crutches~\cite{Melis1999}. 

Walking on uneven ground resulted in longer double support phases and shorter single support phases of the crutches. This further corroborates the conclusion that lateral stability was harder to maintain on the irregular surface. Hence, longer double support phases of the crutches were necessary to maintain balance. This behavior also contributed to the reported increase in impulse of the crutch forces.

\subsection*{Symmetry indices}
In general, on both ground types, compliant knee joints resulted in lower symmetry indices and, therefore, in higher gait symmetry. In chronic stroke patients, asymmetry of spatiotemporal gait parameters was linked to higher risk of falling and to higher balance impairments~\cite{Lewek2014}. This could indicate that rigid knee joints resulted in more difficulty to maintain stability since higher gait asymmetry was observed. Therefore, the higher gait symmetry observed with compliant knee joints may have reduced the physical effort needed by the pilot when walking with the exoskeleton.

\subsection*{Subjective user feedback}
Anecdotally, the user reported that walking in the exoskeleton with rigid knee joint felt much jerkier, especially on uneven ground. He also reported feeling more forced to follow what the exoskeleton commands and having less influence over the walking process with rigid knee joints. Both of these user assessments support our findings. However, it has to be considered that the user was not blinded towards the configuration he was using. 

\subsection*{Limitations and outlook}
The higher walking speed might have increased parameters such as GRF, impulse and base of support. Trials with controlled constant walking speed across conditions could therefore be interesting. This could be achieved by diminishing the step length settings for the compliant knee joints and by triggering the steps with a pre-set timing.

It would be interesting to investigate whether the assumed relationship between crutch force impulse over a gait cycle and effort is valid. Monitoring metabolic cost of walking via gas exchange measurement, as has been performed in other studies evaluating effort during exoskeleton walking~\cite{Asselin2015}, could shed more light on whether this assumption is true. Conducting measurements with force plates to assess load bearing of legs and arms/crutches could further improve our understanding of exoskeleton walking and should thus be considered. However, both additions would increase the complexity of the protocol and the discomfort for the user. Metabolic cost analysis via gas exchange measurements requires relatively long time periods of around 3~min of steady activity to acquire representative results. Force plates have to be specifically targeted by the user during foot placement and can only analyze a single step at a time, which renders acquiring a similar number of steps as presented in this experiment an even more time consuming process. Future investigations should consider recruiting additional users ideally naive to the hardware used in the study. However, study participation requires a high investment of users, as our protocol requires approximately 26 hours of exoskeleton walking sessions, without considering the time needed for preparation and travel.

For this experiment, we left the control, i.e. the PID parameters, fixed between the two conditions. We adjusted the trajectory using a given set of parameters that tweak the exoskeleton movement according to the user's feedback. Specifically, with rigid knee joints, we adapted two trajectory parameters: torso inclination (more hip flexion during double support) and knee extension during stance (lower nominal knee extension during double support). We decided to adjust the latter parameter as we expected knee extension during stance had to be adapted due to the presence or absence of a spring. This was motivated by the fact that a compliant knee joint allows knee flexion under body weight loading. During the training phase with the rigid exoskeleton we discovered that a change in the torso inclination parameter was preferred by the user as he struggled to perform weight shifting from one leg to the other if this setting remained the same as with the compliant exoskeleton. An interesting question for future investigations would be whether the torso inclination parameter could be changed back after initial training, and if differences in inclination and potentially also torso angle observed here between the two knee joint configurations would then disappear. However, it needs to be considered that parameters may always have to be adapted to individual users (for any mode), as they have strong preferences in terms of perceived comfort and performance.

Along the same line, it would be very interesting to compare the same user performing the same protocol with a different exoskeleton for comparison. Performance and walking speed in particular are results of the symbiotic execution of walking that requires synchronized behavior of user and exoskeleton. A longitudinal study with the EksoGT, although with incomplete SCI subjects, found that training time predicted walking speed with an exponential decaying function~\cite{Ramanujam2018}. The reasoning provided is that the double support phase can be reduced quickly over the first few hours of training, but then stagnates. Stride length, on the other hand, was not significantly influenced by the number of training sessions, but well predicted by the participant group (able bodied or incomplete SCI). This suggests that exoskeletons must be improved in design and/or control to achieve faster walking speeds. The  presented results suggest that actuation with compliance may be a step contributing to this goal.

According to our results, compliance was advantageous for the subject in terms of increased performance and decreased effort while walking on both the even and uneven ground. The higher performance and lower effort for the upper body obtained with compliant knees suggest that the compliant exoskeleton may be used by weaker subjects, e.g. due to a higher lesion level, who might not manage to walk with rigid knee joints. Therefore, compliant exoskeletons would be an advantage not only when performing activities of daily living, but also in rehabilitation, because they could potentially allow longer training sessions, might be easier to walk with and could result in more symmetrical and natural gait. The increased performance and decreased effort observed with a compliant exoskeleton support the hypothesis that compliance could also improve wearable robots when walking over uneven ground, similar to what has been demonstrated in underactuated bipedal robots~\cite{Griffin2015} and other legged robots~\cite{Hutter2014}.

This work is important as it will provide exoskeleton developers and clinicians with information about the potential benefits of compliance embedded in powered exoskeletons, paving the way for a novel generation of such assistive devices more robustly coping with real-life environments. In summary, this is the first study investigating the effects of compliant exoskeleton knee joints on uneven ground with a user with motor-complete SCI. Our results suggest that adding compliance to the knee joint is beneficial in all examined domains.

\section*{Conclusion}
This study demonstrated that compliant exoskeleton knees can make gait more efficient and increase walking speed. This was true for both even and uneven ground compared to the rigid configuration. The user could generate higher speeds with less movement, which suggests higher efficiency. Effort of walking was decreased through longer phases of the gait cycle where crutches were not loaded, indicating increased load transfer through the legs. Improved performance and reduced effort might allow the exoskeleton to be used for a longer time and by weaker SCI subjects, making it easier to perform ambulatory tasks in every day life. Thus, we conclude that principles similar to those increasing gait robustness of humanoid and other legged robots are also applicable to powered lower limb exoskeletons. Hence compliance, which could potentially also be added by control, should be considered for exoskeleton designs.
\newpage

\section*{Acknowledgements}
  This work was supported by the Swiss National Science Foundation through the National Center of Competence in Research on Robotics, the ETH Research Grant ETH-22 13-2 and the ETH Zurich Foundation in collaboration with Hocoma AG. We thank Christian Mueller for his advice on statistical analysis. We thank Nöelle Bracher, Philipp Leuenberger and Lucien Rüegg for their help in the data collection process and Werner Witschi for his participation.

\section*{Competing interests}
The authors declare that they have no competing interests.

\section*{Author's contributions}
SOS designed and lead the investigation and wrote the manuscript. GD was responsible for the recording and analysis of the motion capture data and edited the manuscript. CS, CAE, OL and RG supervised the study design and progression, helped with data analysis and interpretation, and worked on the manuscript.

\newpage

%%%%%%%%%%%%%%%%%%%%%%%%%%%%%
% Supplementary Information %
%%%%%%%%%%%%%%%%%%%%%%%%%%%%%
\newpage
%\captionsetup*{format=largeformat}

% Table generated by Excel2LaTeX from sheet 'Descriptive_Latex'
\begingroup
\let\clearpage\relax 
\onecolumn 
\section{Appendix: Statistics} \label{LM}

\pagestyle{empty}
%\begin{landscape}
\begin{table}[htbp]
  \centering
  \footnotesize
  \caption{Descriptive statistics.}
  %\hspace*{-20cm}
  \begin{adjustwidth}{-0.1cm}{}
  %\adjustbox{max width=\textwidth}{
    \begin{tabular}{rlccccccccccc}
    \toprule
          &       & \multicolumn{2}{c}{\textbf{CE}} &       & \multicolumn{2}{c}{\textbf{CU}} &       & \multicolumn{2}{c}{\textbf{RE}} &       & \multicolumn{2}{c}{\textbf{RU}} \\
\cmidrule{3-4}\cmidrule{6-7}\cmidrule{9-10}\cmidrule{12-13}    \multicolumn{1}{l}{\textbf{Parameters}} &       & \textbf{mean ($\mu$)} & \textbf{std} &       & \textbf{mean ($\mu$)} & \textbf{std} &       & \textbf{mean ($\mu$)} & \textbf{std} &       & \textbf{mean ($\mu$)} & \textbf{std} \\
    \midrule
    \multicolumn{1}{l}{\textbf{Spatial and temporal}} &       &       &       &       &       &       &       &       &       &       &       &  \\
    \multicolumn{1}{l}{Cycle duration (s)} &       & 4.053 & 0.145 &       & 4.538 & 0.298 &       & 4.122 & 0.178 &       & 4.358 & 0.217 \\
    \multicolumn{1}{l}{Included angle ($^{\circ}$)} & Left HS  & 18.542 & 1.724 &       & 17.364 & 2.926 &       & 8.093 & 1.304 &       & 7.922 & 1.939 \\
          & Right HS & 20.059 & 1.922 &       & 19.009 & 2.593 &       & 11.156 & 1.313 &       & 11.955 & 1.808 \\
    \multicolumn{1}{l}{Lead angle ($^{\circ}$)} & Left HS  & 0.034 & 1.561 &       & -1.473 & 2.308 &       & -9.206 & 2.071 &       & -10.754 & 1.851 \\
          & Right HS & 0.768 & 1.781 &       & -1.523 & 2.253 &       & -6.472 & 1.683 &       & -8.326 & 1.725 \\
    \multicolumn{1}{l}{Phases duration (s)} & IDS   & 0.471 & 0.086 &       & 0.628 & 0.206 &       & 0.614 & 0.108 &       & 0.727 & 0.171 \\
          & SS    & 1.533 & 0.039 &       & 1.598 & 0.121 &       & 1.426 & 0.049 &       & 1.443 & 0.036 \\
          & FDS   & 0.543 & 0.078 &       & 0.715 & 0.152 &       & 0.550 & 0.130 &       & 0.612 & 0.128 \\
          & SW    & 1.473 & 0.044 &       & 1.564 & 0.097 &       & 1.497 & 0.028 &       & 1.543 & 0.044 \\
    \multicolumn{1}{l}{Phases duration (\%)} & IDS   & 11.584 & 1.743 &       & 13.707 & 3.578 &       & 14.866 & 1.903 &       & 16.562 & 3.021 \\
          & SS    & 38.273 & 1.283 &       & 35.677 & 2.851 &       & 35.056 & 1.510 &       & 33.575 & 1.766 \\
          & FDS   & 13.368 & 1.650 &       & 15.693 & 2.786 &       & 13.292 & 2.541 &       & 13.988 & 2.480 \\
          & SW    & 36.774 & 1.249 &       & 34.923 & 2.430 &       & 36.786 & 1.412 &       & 35.875 & 2.024 \\
    \multicolumn{1}{l}{Speed (m/s)} &       & 0.145 & 0.007 &       & 0.116 & 0.017 &       & 0.100 & 0.008 &       & 0.083 & 0.010 \\
    \multicolumn{1}{l}{Step length (m)} & Left  & 0.298 & 0.022 &       & 0.283 & 0.039 &       & 0.175 & 0.013 &       & 0.163 & 0.020 \\
          & Right & 0.298 & 0.020 &       & 0.267 & 0.029 &       & 0.215 & 0.012 &       & 0.196 & 0.016 \\
    \multicolumn{1}{l}{Stride length (m)} & Left  & 0.596 & 0.019 &       & 0.546 & 0.054 &       & 0.413 & 0.020 &       & 0.363 & 0.029 \\
          & Right & 0.587 & 0.019 &       & 0.546 & 0.052 &       & 0.410 & 0.019 &       & 0.359 & 0.028 \\
    \multicolumn{1}{l}{Trail angle ($^{\circ}$)} & Left HS  & 18.508 & 1.423 &       & 18.837 & 2.175 &       & 17.299 & 1.739 &       & 18.677 & 1.913 \\
          & Right HS & 19.291 & 1.151 &       & 20.531 & 2.565 &       & 17.628 & 1.522 &       & 20.281 & 1.861 \\
    \multicolumn{1}{l}{\textbf{Joint angles and ROMs}} &       &       &       &       &       &       &       &       &       &       &       &  \\
    \multicolumn{1}{l}{Hip ROM ($^{\circ}$)} & Left hip  & 28.133 & 2.284 &       & 30.811 & 1.351 &       & 25.850 & 2.394 &       & 29.322 & 1.708 \\
          & Right hip & 27.712 & 1.193 &       & 29.863 & 1.807 &       & 28.747 & 1.453 &       & 30.070 & 1.851 \\
    \multicolumn{1}{l}{Inclination peak angle ($^{\circ}$)} & Left stance  & 14.298 & 1.243 &       & 16.885 & 1.515 &       & 17.127 & 1.162 &       & 20.362 & 1.226 \\
          & Right stance & 14.538 & 1.249 &       & 16.567 & 1.626 &       & 17.162 & 1.530 &       & 20.363 & 1.404 \\
    \multicolumn{1}{l}{Inclination ROM ($^{\circ}$)} & Left stance  & 7.355 & 1.224 &       & 6.197 & 2.040 &       & 3.700 & 0.724 &       & 3.133 & 1.220 \\
          & Right stance & 7.342 & 1.182 &       & 4.642 & 1.576 &       & 4.330 & 0.898 &       & 3.516 & 1.326 \\
    \multicolumn{1}{l}{Knee ROM ($^{\circ}$)} & Left knee  & 38.938 & 1.859 &       & 41.411 & 2.273 &       & 55.863 & 3.236 &       & 56.034 & 3.257 \\
          & Right knee & 35.970 & 1.484 &       & 37.281 & 2.379 &       & 56.680 & 0.930 &       & 56.757 & 1.717 \\
    \multicolumn{1}{l}{Trunk peak angle ($^{\circ}$)} & Left step  & 19.802 & 1.480 &       & 26.340 & 1.947 &       & 28.194 & 1.603 &       & 34.655 & 1.922 \\
          & Right step & 20.765 & 1.373 &       & 26.540 & 1.676 &       & 30.300 & 1.607 &       & 35.856 & 1.760 \\
    \multicolumn{1}{l}{Trunk ROM ($^{\circ}$)} & Left step  & 3.539 & 0.744 &       & 3.380 & 0.901 &       & 3.692 & 0.835 &       & 3.309 & 0.956 \\
          & Right step & 4.193 & 0.769 &       & 4.477 & 1.107 &       & 6.528 & 1.090 &       & 4.552 & 0.988 \\
    \multicolumn{1}{l}{\textbf{Pelvis trajectory}} &       &       &       &       &       &       &       &       &       &       &       &  \\
    \multicolumn{1}{l}{AP excursion (m)} &       & 0.111 & 0.019 &       & 0.148 & 0.024 &       & 0.136 & 0.019 &       & 0.155 & 0.018 \\
    \multicolumn{1}{l}{ML excursion (m)} &       & 0.135 & 0.031 &       & 0.134 & 0.021 &       & 0.121 & 0.020 &       & 0.118 & 0.023 \\
    \multicolumn{1}{l}{SLAP ratio} &       & 5.530 & 1.039 &       & 3.798 & 0.794 &       & 3.093 & 0.441 &       & 2.371 & 0.301 \\
    \multicolumn{1}{l}{\textbf{Strategy with the crutches}} &       &       &       &       &       &       &       &       &       &       &       &  \\
    \multicolumn{1}{l}{BoS (m2)} & Left stance & 0.274 & 0.022 &       & 0.286 & 0.035 &       & 0.316 & 0.039 &       & 0.346 & 0.026 \\
          & Right stance & 0.282 & 0.026 &       & 0.283 & 0.035 &       & 0.308 & 0.034 &       & 0.344 & 0.025 \\
    \multicolumn{1}{l}{cDS (\%)} &       & 78.468 & 0.607 &       & 81.068 & \multicolumn{1}{r}{1.828} &       & \multicolumn{1}{r}{78.709} & \multicolumn{1}{r}{1.373} &       & \multicolumn{1}{r}{80.315} & 0.927 \\
    \multicolumn{1}{l}{cSS (\%)} &       & 21.532 & 0.607 &       & 18.932 & \multicolumn{1}{r}{1.828} &       & \multicolumn{1}{r}{21.291} & \multicolumn{1}{r}{1.373} &       & \multicolumn{1}{r}{19.685} & 0.927 \\
    \multicolumn{1}{l}{\textbf{Crutch GRFs and impulse}} &       &       &       &       &       &       &       &       &       &       &       &  \\
    \multicolumn{1}{l}{Peak net total force (N)} &       & 503.760 & 37.442 &       & 648.428 & 81.270 &       & 518.711 & 44.111 &       & 654.419 & 41.454 \\
    \multicolumn{1}{l}{Impulse: total support (Ns)} &       & 1291.633 & 125.690 &       & 1842.209 & 276.283 &       & 1566.003 & 124.765 &       & 1996.118 & 170.724 \\
    \multicolumn{1}{l}{Impulse: left crutch (Ns)} & AP    & 226.452 & 33.553 &       & 241.218 & 34.493 &       & 230.742 & 27.802 &       & 202.403 & 28.871 \\
          & ML    & 172.026 & 44.088 &       & 313.201 & 64.305 &       & 189.039 & 40.167 &       & 335.900 & 49.600 \\
          & Vertical     & 606.904 & 104.256 &       & 874.227 & 155.946 &       & 788.467 & 91.025 &       & 977.661 & 118.825 \\
    \multicolumn{1}{l}{Impulse: right crutch (Ns)} & AP    & 204.736 & 36.227 &       & 223.888 & 49.333 &       & 228.326 & 33.916 &       & 243.563 & 32.287 \\
          & ML    & 184.584 & 30.155 &       & 319.562 & 62.369 &       & 212.426 & 60.471 &       & 339.252 & 46.594 \\
          & Vertical     & 547.958 & 78.186 &       & 784.460 & 143.801 &       & 643.771 & 79.630 &       & 839.027 & 84.974 \\
    \bottomrule
    \end{tabular}%
   % }
   \end{adjustwidth}
    %\hspace*{-20cm}
  \label{tab:descriptive_statistics}%
\end{table}%
%\end{landscape}
\pagestyle{plain}

\newpage
% Table generated by Excel2LaTeX from sheet 'LM_all'
\begin{table}[htbp]
  \centering
  \footnotesize
  \caption{Results from the linear regression models. Non-significant p-values are reported in bold.}
  %\hspace*{-3.8cm}
  \begin{adjustwidth}{-1.1cm}{}
    \begin{tabular}{lrcccccccccccccc}
    \toprule
          &       & \multicolumn{5}{c}{\textbf{Ground effect}} &       & \multicolumn{5}{c}{\textbf{Configuration effect  }} &       &       &  \\
\cmidrule{3-7}\cmidrule{9-13}          &       & \multicolumn{2}{c}{\textbf{Rigid leg}} &       & \multicolumn{2}{c}{\textbf{Soft leg}} &       & \multicolumn{2}{c}{\textbf{Even ground}} &       & \multicolumn{2}{c}{\textbf{Uneven ground}} &       & \multicolumn{2}{c}{\textbf{Interaction effect }} \\
\cmidrule{3-4}\cmidrule{6-7}\cmidrule{9-10}\cmidrule{12-13}\cmidrule{15-16}    \textbf{Parameter} &       & \textbf{t-value} & \textbf{p-value} &       & \textbf{t-value} & \textbf{p-value} &       & \textbf{t-value} & \textbf{p-value} &       & \textbf{t-value} & \textbf{p-value} &       & \textbf{t-value} & \textbf{p-value} \\
    \midrule
    \textbf{Spatial and temporal} &       &       &       &       &       &       &       &       &       &       &       &       &       &       &  \\
    Cycle duration &       & 10.2973 & 0.0000 &       & -16.3799 & 0.0000 &       & -3.5632 & 0.0004 &       & -5.6208 & 0.0000 &       & 6.6537 & 0.0000 \\
    Included angle  & \multicolumn{1}{l}{Left} & -0.8072 & \textbf{0.4199} &       & 4.3142 & 0.0000 &       & 58.4122 & 0.0000 &       & -31.9837 & 0.0000 &       & -2.9182 & 0.0037 \\
          & \multicolumn{1}{l}{Right} & 3.8086 & 0.0002 &       & 3.8731 & 0.0001 &       & 50.1048 & 0.0000 &       & -24.0550 & 0.0000 &       & -5.3939 & 0.0000 \\
    Lead angle & \multicolumn{1}{l}{Left} & -6.7763 & 0.0000 &       & 5.1059 & 0.0000 &       & 47.7647 & 0.0000 &       & -29.0706 & 0.0000 &       & 0.1084 & \textbf{0.9137} \\
          & \multicolumn{1}{l}{Right} & -8.7178 & 0.0000 &       & 8.3320 & 0.0000 &       & 40.1901 & 0.0000 &       & -22.8864 & 0.0000 &       & -1.2551 & \textbf{0.2100} \\
    Phases duration (s) & \multicolumn{1}{l}{IDS} & 7.3004 & 0.0000 &       & -7.9036 & 0.0000 &       & -11.0189 & 0.0000 &       & 4.5915 & 0.0000 &       & 1.7837 & \textbf{0.0750} \\
          & \multicolumn{1}{l}{SS} & 2.4987 & 0.0128 &       & -7.4971 & 0.0000 &       & 18.8585 & 0.0000 &       & -16.5731 & 0.0000 &       & 4.4005 & 0.0000 \\
          & \multicolumn{1}{l}{FDS} & 4.3346 & 0.0000 &       & -9.3864 & 0.0000 &       & -0.6402 & \textbf{0.5223} &       & -5.1910 & 0.0000 &       & 4.7713 & 0.0000 \\
          & \multicolumn{1}{l}{SW} & 8.1168 & 0.0000 &       & -12.5756 & 0.0000 &       & -5.2283 & 0.0000 &       & -2.6538 & 0.0082 &       & 4.9791 & 0.0000 \\
    Phases duration (\%) & \multicolumn{1}{l}{IDS} & 6.1543 & 0.0000 &       & -5.9590 & 0.0000 &       & -14.0572 & 0.0000 &       & 7.4109 & 0.0000 &       & 0.9470 & \textbf{0.3440} \\
          & \multicolumn{1}{l}{SS} & -7.4009 & 0.0000 &       & 10.0374 & 0.0000 &       & 18.9833 & 0.0000 &       & -7.5164 & 0.0000 &       & -3.4098 & 0.0007 \\
          & \multicolumn{1}{l}{FDS} & 2.5130 & 0.0123 &       & -6.4954 & 0.0000 &       & 0.3244 & \textbf{0.7457} &       & -4.4054 & 0.0000 &       & 3.5995 & 0.0003 \\
          & \multicolumn{1}{l}{SW} & -4.7139 & 0.0000 &       & 7.4106 & 0.0000 &       & -0.0726 & \textbf{0.9421} &       & 3.5243 & 0.0005 &       & -2.9765 & 0.0030 \\
    Speed &       & -15.2543 & 0.0000 &       & 20.4671 & 0.0000 &       & 48.0208 & 0.0000 &       & -21.0861 & 0.0000 &       & -6.8529 & 0.0000 \\
    Step length & \multicolumn{1}{l}{Left} & -4.7417 & 0.0000 &       & 4.6340 & 0.0000 &       & 57.7527 & 0.0000 &       & -34.1036 & 0.0000 &       & -0.7634 & \textbf{0.4455} \\
          & \multicolumn{1}{l}{Right} & -9.1755 & 0.0000 &       & 11.6743 & 0.0000 &       & 47.2873 & 0.0000 &       & -24.4237 & 0.0000 &       & -3.6184 & 0.0003 \\
    Stride length & \multicolumn{1}{l}{Left } & -15.6889 & 0.0000 &       & 12.0399 & 0.0000 &       & 67.3534 & 0.0000 &       & -40.9054 & 0.0000 &       & 0.0781 & \textbf{0.9378} \\
          & \multicolumn{1}{l}{Right} & -15.2543 & 0.0000 &       & 20.4671 & 0.0000 &       & 48.0208 & 0.0000 &       & -21.0861 & 0.0000 &       & -6.8529 & 0.0000 \\
    Trail angle & \multicolumn{1}{l}{Left } & 6.6667 & 0.0000 &       & -1.2347 & \textbf{0.2175} &       & 6.9063 & 0.0000 &       & -0.5566 & \textbf{0.5780} &       & -3.1031 & 0.0020 \\
          & \multicolumn{1}{l}{Right} & 13.6882 & 0.0000 &       & -4.9504 & 0.0000 &       & 10.1328 & 0.0000 &       & -0.9240 & \textbf{0.3559} &       & -4.4611 & 0.0000 \\
    \textbf{Joint angles and ROMs} &       &       &       &       &       &       &       &       &       &       &       &       &       &       &  \\
    Hip ROM & \multicolumn{1}{l}{Left} & 13.4598 & 0.0000 &       & -8.0331 & 0.0000 &       & 10.4511 & 0.0000 &       & -4.1311 & 0.0000 &       & -1.8831 & \textbf{0.0602} \\
          & \multicolumn{1}{l}{Right} & -4.7417 & 0.0000 &       & 4.6340 & 0.0000 &       & 57.7527 & 0.0000 &       & -34.1036 & 0.0000 &       & -0.7634 & \textbf{0.4455} \\
    Inclination peak angle & \multicolumn{1}{l}{Left} & 21.9368 & 0.0000 &       & -13.5691 & 0.0000 &       & -22.6455 & 0.0000 &       & 16.8702 & 0.0000 &       & -2.6921 & 0.0073 \\
          & \multicolumn{1}{l}{Right} & 18.6816 & 0.0000 &       & -9.1620 & 0.0000 &       & -18.0773 & 0.0000 &       & 15.8485 & 0.0000 &       & -4.1857 & 0.0000 \\
    Inclination ROM & \multicolumn{1}{l}{Left} & -4.0859 & 0.0001 &       & 6.4575 & 0.0000 &       & 31.1069 & 0.0000 &       & -15.8015 & 0.0000 &       & -2.6070 & 0.0094 \\
          & \multicolumn{1}{l}{Right} & -5.9301 & 0.0000 &       & 15.2298 & 0.0000 &       & 25.9272 & 0.0000 &       & -5.8698 & 0.0000 &       & -8.4166 & 0.0000 \\
    Knee ROM & \multicolumn{1}{l}{Left} & 0.5111 & \textbf{0.6095} &       & -5.7342 & 0.0000 &       & -59.8949 & 0.0000 &       & 31.3571 & 0.0000 &       & 4.2225 & 0.0000 \\
          & \multicolumn{1}{l}{Right} & -9.1755 & 0.0000 &       & 11.6743 & 0.0000 &       & 47.2873 & 0.0000 &       & -24.4237 & 0.0000 &       & -3.6184 & 0.0003 \\
    Trunk peak angle & \multicolumn{1}{l}{Left} & 32.5127 & 0.0000 &       & -25.4561 & 0.0000 &       & -49.8676 & 0.0000 &       & 29.9402 & 0.0000 &       & 0.2377 & \textbf{0.8122} \\
          & \multicolumn{1}{l}{Right} & 29.5587 & 0.0000 &       & -23.7768 & 0.0000 &       & -59.9065 & 0.0000 &       & 35.4629 & 0.0000 &       & 0.7171 & \textbf{0.4736} \\
    Trunk ROM & \multicolumn{1}{l}{Left} & -3.8371 & 0.0001 &       & 1.2296 & \textbf{0.2194} &       & -1.8141 & 0.0702 &       & -0.5092 & \textbf{0.6108} &       & 1.3756 & \textbf{0.1695} \\
          & \multicolumn{1}{l}{Right} & -16.8066 & 0.0000 &       & -1.8692 & \textbf{0.0621} &       & -23.4563 & 0.0000 &       & 0.4597 & \textbf{0.6459} &       & 11.7629 & 0.0000 \\
    \textbf{Pelvis trajectory} &       &       &       &       &       &       &       &       &       &       &       &       &       &       &  \\
    AP excursion  &       & 8.1142 & 0.0000 &       & -12.3459 & 0.0000 &       & -12.7619 & 0.0000 &       & 2.1219 & 0.0343 &       & 4.7991 & 0.0000 \\
    ML excursion  &       & -0.9074 & \textbf{0.3646} &       & 0.4337 & \textbf{0.6646} &       & 5.7023 & 0.0000 &       & -3.7036 & 0.0002 &       & 0.2122 & \textbf{0.8320} \\
    SLAP\_ratio &       & -8.6662 & 0.0000 &       & 16.0832 & 0.0000 &       & 34.5384 & 0.0000 &       & -12.2565 & 0.0000 &       & -7.4171 & 0.0000 \\
    \textbf{Strategy with the crutches} &       &       &       &       &       &       &       &       &       &       &       &       &       &       &  \\
    BoS   & \multicolumn{1}{l}{Left} & 7.9000 & 0.0000 &       & -2.4021 & 0.0166 &       & -13.0350 & 0.0000 &       & 11.3299 & 0.0000 &       & -2.9344 & 0.0035 \\
          & \multicolumn{1}{l}{Right} & 9.8310 & 0.0000 &       & -0.2268 & \textbf{0.8207} &       & -8.7411 & 0.0000 &       & 12.1214 & 0.0000 &       & -5.8366 & 0.0000 \\
    cDS   &       & -15.2543 & 0.0000 &       & 20.4671 & 0.0000 &       & 48.0208 & 0.0000 &       & -21.0861 & 0.0000 &       & -6.8529 & 0.0000 \\
    cSS   &       & -9.1755 & 0.0000 &       & 11.6743 & 0.0000 &       & 47.2873 & 0.0000 &       & -24.4237 & 0.0000 &       & -3.6184 & 0.0003 \\
    \textbf{Crutch GRFs and impulse} &       &       &       &       &       &       &       &       &       &       &       &       &       &       &  \\
    Peak\_GRF\_net\_total &       & 24.2655 & 0.0000 &       & -20.0141 & 0.0000 &       & -3.1567 & 0.0017 &       & 0.7666 & \textbf{0.4437} &       & 0.9803 & \textbf{0.3273} \\
    Impulse: total support &       & 23.2385 & 0.0000 &       & -23.0156 & 0.0000 &       & -17.5038 & 0.0000 &       & 5.9499 & 0.0000 &       & 3.9827 & 0.0001 \\
    Impulse: left crutch  & \multicolumn{1}{l}{AP} & -7.7842 & 0.0000 &       & -3.1382 & 0.0018 &       & -1.3915 & \textbf{0.1646} &       & -7.6288 & 0.0000 &       & 7.2455 & 0.0000 \\
          & \multicolumn{1}{l}{ML} & 26.7784 & 0.0000 &       & -19.9167 & 0.0000 &       & -3.6629 & 0.0003 &       & 2.9615 & 0.0032 &       & -0.6345 & \textbf{0.5260} \\
          & \multicolumn{1}{l}{V} & 14.6645 & 0.0000 &       & -16.0316 & 0.0000 &       & -16.6172 & 0.0000 &       & 5.7365 & 0.0000 &       & 3.7057 & 0.0002 \\
    Impulse: right crutch  & \multicolumn{1}{l}{AP} & 3.5349 & 0.0004 &       & -3.4377 & 0.0006 &       & -6.4621 & 0.0000 &       & 3.2660 & 0.0012 &       & 0.5558 & \textbf{0.5786} \\
          & \multicolumn{1}{l}{ML} & 20.9631 & 0.0000 &       & -17.2620 & 0.0000 &       & -5.4341 & 0.0000 &       & 2.3287 & 0.0202 &       & 0.8245 & \textbf{0.4100} \\
          & \multicolumn{1}{l}{V} & 18.4852 & 0.0000 &       & -17.3236 & 0.0000 &       & -10.7108 & 0.0000 &       & 3.6963 & 0.0002 &       & 2.3896 & 0.0172 \\
    \textbf{Symmetry indices} &       &       &       &       &       &       &       &       &       &       &       &       &       &       &  \\
    Double support phase &       & 1.7515 & \textbf{0.0804} &       & 0.0025 & \textbf{0.9980} &       & -11.6078 & 0.0000 &       & 8.2894 & 0.0000 &       & -1.0737 & \textbf{0.2834} \\
    Included angle &       & -3.1478 & 0.0017 &       & 0.4724 & \textbf{0.6368} &       & 9.9125 & 0.0000 &       & -7.8221 & 0.0000 &       & 1.5527 & \textbf{0.1211} \\
    Single support phase &       & -3.5931 & 0.0004 &       & 3.0553 & 0.0024 &       & 21.2155 & 0.0000 &       & -12.6014 & 0.0000 &       & -0.2177 & \textbf{0.8277} \\
    Step length &       & 1.2158 & \textbf{0.2246} &       & -2.5573 & 0.0108 &       & 15.2074 & 0.0000 &       & -10.7102 & 0.0000 &       & 1.2786 & \textbf{0.2016} \\
    \bottomrule
    \end{tabular}%
    \end{adjustwidth}
    %\hspace*{-3.8cm}
  \label{tab:LM_results}%
\end{table}%

\endgroup

\newpage
\twocolumn
\bibliographystyle{IEEEtran}
\bibliography{IEEEabrv,article}

\end{document}